\pdfoutput=1

\documentclass[11pt]{article}

\usepackage[]{EMNLP2023}

\usepackage{times}
\usepackage{latexsym}

\usepackage[T1]{fontenc}

\usepackage[utf8]{inputenc}

\usepackage{microtype}

\usepackage{booktabs,multirow}

\usepackage{inconsolata}

\usepackage{xspace}
\usepackage{float}
\usepackage{array}
\usepackage{booktabs}
\usepackage{verbatim}
\usepackage{lipsum}  
\usepackage{subcaption}
\usepackage{rotating}
\usepackage{multirow}    
\usepackage{colortbl}
\usepackage{booktabs}
\usepackage{tikz}
\usepackage{tikz-qtree}
\newcolumntype{L}[1]{>{\raggedright\let\newline\\\arraybackslash\hspace{0pt}}m{#1}}
\newcolumntype{C}[1]{>{\centering\let\newline\\\arraybackslash\hspace{0pt}}m{#1}}
\newcolumntype{R}[1]{>{\raggedleft\let\newline\\\arraybackslash\hspace{0pt}}m{#1}}

\newcommand{\modelName}{ChapGTP\xspace} 

\title{\modelName, ILLC's Attempt at Raising a BabyLM:\\Improving Data Efficiency by Automatic Task Formation}

\author{{Jaap Jumelet} \\ \textbf{Anna Langedijk} \And {Michael Hanna} \\  \textbf{Charlotte Pouw} \\[5pt]
Institute for Logic, Language and Computation (ILLC) \\University of Amsterdam \And {Marianne de Heer Kloots} \\ \textbf{Oskar van der Wal}}

\begin{document}
\maketitle

\begin{abstract}
We present the submission of the ILLC at the University of Amsterdam to the BabyLM challenge \citep{warstadt-et-al-2023-babylm}, in the \texttt{strict-small} track.
Our final model, \modelName, is a masked language model that was trained for 200 epochs, aided by a novel data augmentation technique called Automatic Task Formation.
We discuss in detail the performance of this model on the three evaluation suites: BLiMP, (Super)GLUE, and MSGS.
Furthermore, we present a wide range of methods that were ultimately not included in the model, but may serve as inspiration for training LMs in low-resource settings.
\end{abstract}

\section{Introduction}
Modern language models (LMs) are trained on datasets that are many orders of magnitude larger than the amount of text a human can read in a single lifetime.
Driven by the \textit{scaling law} paradigm, which states that model performance scales as a power law with model and data size, language model training has become increasingly data hungry \citep{DBLP:journals/corr/abs-2001-08361, hoffmann2022empirical}.
This has raised questions about the \textit{efficiency} of the paradigm: is it possible to train proficient models on amounts of data similar to that what humans process when learning language?
The \textbf{BabyLM} challenge \citep{warstadt-et-al-2023-babylm} proposes a community effort to find efficient training strategies for model training, providing a fixed, ``developmentally plausible'' training data set.

This paper presents the submission of the Institute of Logic, Language and Computation at the University of Amsterdam to the BabyLM challenge.
We participated in the \texttt{strict-small} track of the challenge, which limits the amount of training data to a fixed set of 10 million tokens.
The usage of any sources trained on external data was not allowed, which forced us to utilize the training data as efficiently as possible.
Evaluation is based on various benchmarks, including BLiMP \citep{warstadt-etal-2020-blimp-benchmark}, (Super)GLUE \citep{wang-etal-2018-glue, superglue}, and MSGS \citep{warstadt-etal-2020-learning}.

Our final model, \textbf{\modelName}\footnote{\textbf{Chap}eroned \textbf{G}eneralised \textbf{T}ask formation and \textbf{P}retraining, DynaBench ID 1448. HuggingFace hub link: \url{https://huggingface.co/mwhanna/ChapGTP}}, is a bidirectional masked LM based on the \texttt{DeBERTa} architecture \citep{DBLP:conf/iclr/HeGC23}.
Our core contribution is a novel data augmentation technique called \textbf{Automatic Task Formation (ATF)}, which generates meaningful textual formulations from the existing training data based on pre-defined templates. These formulations are tailored for learning specific tasks such as question answering and sentiment classification. 
The procedure relies solely on shallow surface heuristics, and requires no external data or expert labeling.

Besides ATF, we explored many other strategies: prosodic guidance, formal languages, tokenizer and model engineering, emergent language games, and grokking. Although not all of these were included in \modelName, many showed potential. Notably, we find that ``pre-pre-training'' a language model on constituency-labeled text (induced by an unsupervised constituency parser) or on synthetic emergent languages (generated by neural agents in a referential game with real images) can lead to improvements on the final evaluation benchmarks---but more research is needed to explore the practicality and effectiveness of these approaches in more detail. We hope that our discussion of the various strategies for training data-efficient language models will inspire other researchers and engineers working on NLP in low-resource settings.

\section{Data-efficient NLP}
The exponential growth in computing resources needed to train recent language models has underscored the need for more data-efficient models. Increased model training efficiency would avoid environmental harms \citep{DBLP:journals/cacm/SchwartzDSE20} and ensure the model openness and accountability that is needed to democratize technological development \citep{ahmed_-democratization_2020, liesenfeld_opening_2023}. From a cognitive perspective, which aims to model human-like generalization abilities, sample efficiency should be more of priority than is currently reflected in leaderboard-like model comparisons \citep{linzen_how_2020}.


Language models' resource consumption can be decreased at all stages of model development, on both the model and data sides; see \citet{DBLP:conf/iclr/HeGC23} for an overview. On the modeling side, many studies have aimed to improve data-efficiency by injecting neural models with inductive biases that aid generalization. Examples of such work include distilling inductive biases from other neural models \citep{abnar_transferring_2020} or Bayesian learning algorithms \citep{mccoy_modeling_2023}. Other work has compared different types of bias by transfer learning to English after ``pre-pre-training'' models on synthetically generated structures \citep{papadimitriou_pretrain_2023}. 

Most relevant to the BabyLM challenge is \citeposs{huebner-etal-2021-babyberta} work inspired by child language learning abilities, which drastically decreased model parameters as well as training data size. They pre-trained RoBERTa-base from scratch on a developmentally plausible amount of data, resulting in a model with lower grammatical competence than the original, large-scale model \citep{liu2019roberta}. However, via careful hyperparameter tuning, they developed \textit{BabyBERTa}, which performs well even with acquisition-scale training data. Their model has only 8 million parameters, 8912 vocabulary items and---importantly---does not predict unmasked tokens.

Data-oriented approaches provide a complementary strategy for improving training efficiency. One successful strategy is to \emph{filter} the training data, for example by removing duplicates \citep{lee_deduplicating_2022}, or excluding thematic document clusters that lead to undesirable model behavior \citep{kaddour2023minipile}. \citet{mishra-sachdeva-2020-need} used human-inspired heuristics to remove irrelevant and redundant data, aiming to select the optimal dataset for learning a specific task. Via a combination of coarse and fine pruning techniques, they achieved competitive results on out-of-distribution NLI datasets with only $\sim$2\% of the SNLI training set.

Finally, \emph{data augmentation} has proven to be useful in low-resource settings. Such techniques aim to diversify the set of training examples without collecting more data \citep{feng_survey_2021}; this can lead to task-specific or domain-general improvement on model performance. \citet{fabbri_template-based_2020} showed that performance on a downstream question answering (QA) task increased when models' training data was augmented with synthetically generated questions that helped models learn more complex question-context relationships. Their most successful approach used simple templates to generate wh-questions based on sentences retrieved from the original training data.

\citet{jia_question_2022} showed that including automatically generated question-answer pairs in pre-training data leads to a better encoding of contextual information in token-level representations. They found that this \emph{question-infused pre-training} strategy results in improved model performance on a range of standard NLP tasks beyond QA, including paraphrase detection, named entity recognition, and sentiment analysis. 
\section{The BabyLM Challenge}

The BabyLM Challenge is a shared task that challenges researchers to train a language model from scratch on an amount of linguistic data similar to what is available to a child. The task has two main goals: 1) developing novel techniques for learning efficiently in low-resource settings; and 2) increasing access to cognitively plausible models of language, which could improve our understanding of human language learning.

\paragraph{Training Data}
The BabyLM Challenge offers a \emph{developmentally plausible} training dataset, drawing inspiration from the linguistic input children typically receive until the age of 13. The dataset contains fewer than 100 million words and predominantly uses transcribed speech, as children are primarily exposed to spoken language during their early years. The data come from various domains: \emph{child-directed speech} (CHILDES; \citealp{macwhinney2000childes}), \emph{dialogue} (Switchboard Dialog Act Corpus; \citealp{stolcke2000dialogue}), \emph{subtitles} (OpenSubtitles, \citealp{lison2016opensubtitles2016}, and QCRI Educational Domain Corpus (QED), \citealp{abdelali2014amara}), \emph{simple written English} (Simple Wikipedia, Children's Book Test \citealp{hill2015goldilocks}, Children Stories Text Corpus), and \emph{regular written English} (Wikipedia, Standardized Project Gutenberg Corpus \citealp{DBLP:journals/corr/abs-1812-08092}).


The challenge features three participation tracks: \texttt{strict}, \texttt{strict-small}, and \texttt{loose}. In the \texttt{strict} track, the training dataset is limited to 100 million written words extracted from the sources above. In the \texttt{strict-small} track, the training dataset is further restricted to a subset of merely 10 million words from the \texttt{strict} dataset. In the \texttt{loose} track, models could additionally be trained on an unlimited amount of non-linguistic data (e.g. symbolic data, audio, images, etc.).
For the exact number and proportion of words per data source included in the \texttt{strict} and \texttt{strict-small} dataset, see \citet{warstadt-et-al-2023-babylm}.

\paragraph{Evaluation}

The evaluation of BabyLM models is based on various benchmarks, namely BLiMP \citep{warstadt-etal-2020-blimp-benchmark}, (Super)GLUE \citep{wang-etal-2018-glue, superglue}, and MSGS \citep{warstadt-etal-2020-learning}. These benchmarks cover a wide range of linguistic phenomena and aim to collectively provide a comprehensive assessment of a model's linguistic capabilities. BabyLM provides filtered versions of the benchmarks, where each example only includes words that have appeared in the \texttt{strict-small} training set at least twice.

BLiMP (Benchmark of Linguistic Minimal Pairs for English) targets linguistic acceptability judgments, and contains sentence pairs that differ in grammatical acceptability based on only one distinct linguistic element. The sentence pairs cover 12 phenomena from English morphology, syntax and semantics, such as anaphor agreement, binding and filler-gap constructions. If a language model is sensitive to the linguistic phenomenon under consideration, it should assign higher probability to the acceptable sentence of the minimal pair.

GLUE (General Language Understanding Evaluation) is a collection of diverse natural language understanding tasks, such as sentiment analysis and textual entailment. SuperGLUE is an improvement upon GLUE and additionally includes coreference resolution and question answering tasks. Both GLUE and SuperGLUE are used for BabyLM evaluation, summing to 11 tasks in total.

MSGS (Mixed Signals Generalization Set) aims to test whether a model prefers linguistic or surface generalizations, through a range of binary classification tasks. It contains \textit{unambiguous tasks} that can be solved by relying on either a surface \textit{or} a linguistic feature (not both), and \textit{ambiguous tasks} that can be solved both by relying on a surface feature \textit{and} by relying on a linguistic feature. The unambiguous tasks test whether a model represents the features of interest in the first place. The ambiguous tasks tests the model's preference for linguistic or surface generalization. The BabyLM evaluation includes 5 unambiguous tasks and 6 ambiguous tasks.

Evaluation on BLiMP is performed in a zero-shot setting, by calculating the proportion of minimal pairs for which the model assigns higher probability to the acceptable sentence. For (Super)GLUE and MSGS, evaluation involves fine-tuning models on each task and then calculating accuracy or macro-F1. The task-specific scores are averaged to arrive at a final score for each of the three benchmarks.
\section{\modelName}
In this section we describe the components of our final model, \modelName, that we submitted to the \texttt{strict-small} track of BabyLM.
The results of the model are presented in \S\ref{sec:results}.
In \S\ref{sec:did_not_work} we describe various approaches that were not successful, but that may inspire future work on improving data efficiency in language modeling.

\paragraph{Model Architecture}
In our experiments we initially considered both causal and masked LM architectures; we ultimately chose a masked LM since it outperformed causal LMs on all evaluation tasks.
The model is based on the \texttt{DeBERTa-small} architecture \citep{DBLP:conf/iclr/HeGC23}: a 6 layer bidirectional transformer, 12 attention heads, a hidden state size of 768, and intermediate state size of 3072.
The final model has 43.5 million parameters.

\paragraph{Data Processing}
We use a Byte-Pair Encoding tokenizer \citep{sennrich-etal-2016-neural}, which we train on the \texttt{strict-small} corpora, limited to a vocabulary size of 10,000 tokens.
This relatively small vocabulary size was sufficient for the challenge, and allowed for more compact models and faster model training.

We preprocessed the corpora by appending all sentences together, separated by a special separator token.
This ensures that consecutive sentences within a paragraph will occur together in a single batch item, allowing the model to leverage inter-sentential information.
It also significantly improves training speed, since all batches are fully filled up, with little to no padding overhead.

\paragraph{Model Training}
We train the model with a masked token prediction objective, with a token masking probability of 15\%.
We train for 200 epochs with a batch size of 64 and a maximum sentence length of 128.
We investigate the impact of the number of epochs in more detail in \S\ref{sec:results}.
We use the \texttt{AdamW} optimizer \citep{DBLP:conf/iclr/LoshchilovH19}, with a cosine learning rate scheduler that interpolates from $5\cdot 10^{-4}$ to 0, weight decay set to 0.1, and gradient accumulation for 8 steps.
We train models using the \texttt{transformers} library \citep{wolf-etal-2020-transformers}.

\section{ATF: Automatic Task Formation}\label{sec:atf}
The \texttt{strict-small} track of the BabyLM challenge did not permit the usage of external data sources to improve the learning procedure.
It was therefore vital to use all data in the training corpora as efficiently as possible. To this end, we defined \textbf{Automatic Task Formation (ATF)}, a procedure that looks for simple \texttt{regex} patterns in the training data that we can use to augment the data.
The main goal of ATF was to improve performance on the GLUE tasks: we hoped that if the training data were augmented with patterns that resembled data found in GLUE, the model could already start learning representations useful for GLUE tasks during pre-training.

\paragraph{Question Answering}
The text in the pre-training corpora already contains questions, such as those found in dialogue. 
However, most of these questions do not require a retrieval-based approach of finding the answer to the question (e.g. ``\textit{How are you doing?}'').
To aid the model with retrieval-based question answering, which is vital for GLUE tasks like QNLI \citep{rajpurkar-etal-2018-know}, we augment the training corpus with question-answer pairs about various topics.
The patterns we consider are:
\begin{enumerate}
    \item \textbf{Birth date}\\
    The (Simple) Wikipedia data contains many patterns of the form `\textit{$\langle$Name$\rangle$ (born $\langle$DD$\rangle$ $\langle$Month$\rangle$ $\langle$YYYY$\rangle$)}'.
    For each such instance, we add a question-answer pair of the form `\textit{On what date was $\langle$Name$\rangle$ born?} \texttt{[SEP]} \textit{$\langle$DD$\rangle$ $\langle$Month$\rangle$ $\langle$YYYY$\rangle$}'.
    \item \textbf{Nationality \& Profession}\\
    The Simple Wikipedia articles describe people in the same template: `\textit{$\langle$Name$\rangle$ (born X) is a $\langle$Profession$\rangle$ from $\langle$Nationality$\rangle$}'.
    We use this pattern to augment the data with question-answer pairs of the form `\textit{Where is $\langle$Name$\rangle$ from?}' and `\textit{What is the profession of $\langle$Name$\rangle$?}'.
    \item \textbf{Discovery, Founding \& Naming}\\
    We consider three other patterns, of the form `\textit{$\langle$Name$\rangle$ was discovered in $\langle$Year$\rangle$}', `\textit{$\langle$Name$\rangle$ was founded in $\langle$Year$\rangle$}', and `\textit{$\langle$Name$_1\rangle$ was named after $\langle$Name$_2\rangle$}'.
\end{enumerate}
In total this procedure yielded 1663 question-answer pairs that we append to the training corpus.

\paragraph{Sentiment Classification}
To aid the model with the sentiment classification task of SST-2 \citep{socher-etal-2013-recursive}, we augment our dataset by exploiting sentences containing sentiment carrying tokens.
After each sentence that contains a token from a list of positive tokens (\textit{great, terrific, etc.}) or negative tokens (\textit{not good, terrible, etc.})\footnote{We report the full lists in Appendix \ref{sec:app_sentiment}.}, we add a special sentiment token followed by the sentence sentiment.
Sentence sentiment is solely based on the presence of a positive or negative token; we skip sentences containing both positive and negative tokens.
The procedure yielded 2500 positive and 2500 negative sentences, which we appended to the training corpus.

Note that we do not modify the masked language modeling training objective for this: the prediction of answers (as well as questions) is performed in the same way as any other token prediction.
Incorporating the procedure with a separate classification head is something that we leave open for future work.

\section{Results}\label{sec:results}

\begin{figure*}
    \centering
    \includegraphics[width=0.24\textwidth,trim={0 -2.6cm 0 0}]{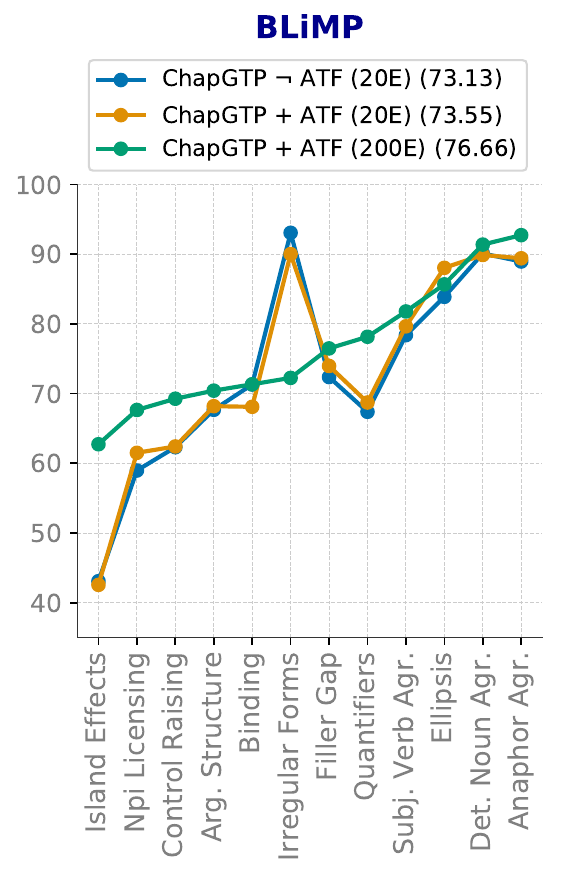}
    \hfill
    \includegraphics[width=0.24\textwidth,trim={0 -5.15cm 0 0}]{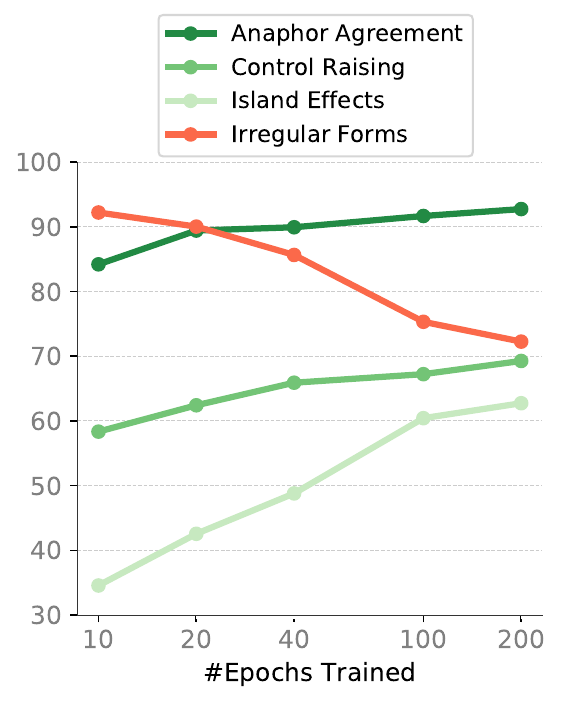}
    \hfill
    \includegraphics[width=0.24\textwidth,trim={0 -4.05cm 0 0}]{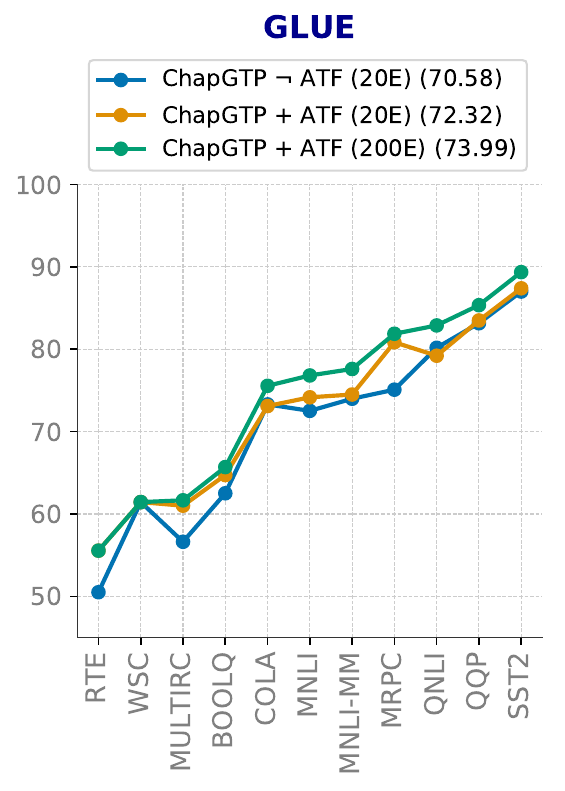}
    \hfill
    \includegraphics[width=0.24\textwidth]{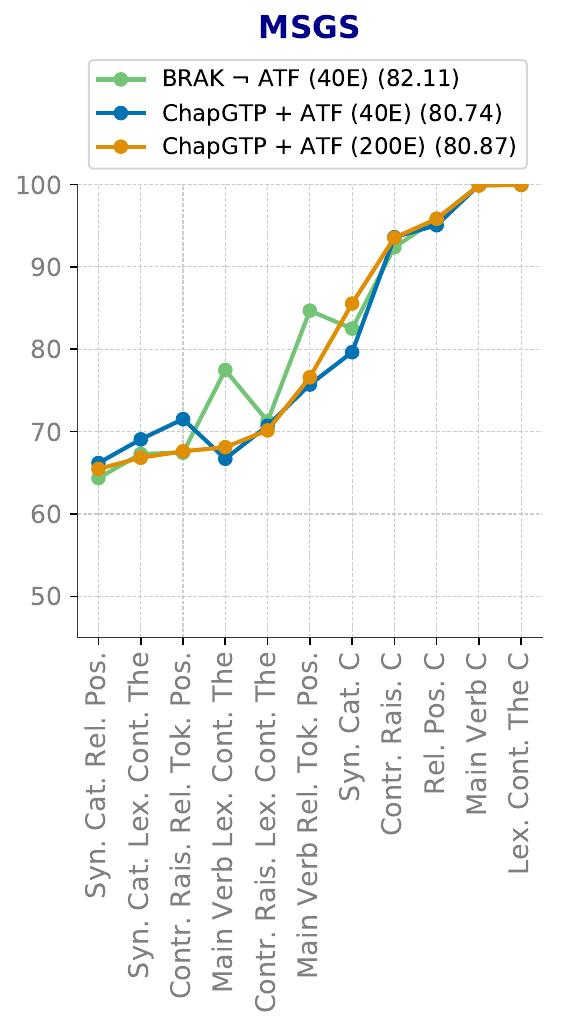}
    \caption{
    (A) Results for BLiMP on the individual conditions, ordered increasingly by the performance of the final 200\textsc{e} model.
    (B) Inverse scaling behavior on the Irregular Forms condition, which worsens as the amount of training is increased. For other tasks the opposite is true: training for longer leads to a monotonic improvement.
    (C) Results for the individual GLUE tasks, ordered similarly to the BLiMP scores in (A).
    (D) Results for the individual MSGS tasks, including the BRAK model that outperforms the \modelName on average.
    }
    \label{fig:blimp}
\end{figure*}

\begin{table}[]
    {\sf\footnotesize
    \centering
    \setlength\extrarowheight{3pt} 
    \begin{tabular*}{7.7cm}{p{2.6cm}@{} @{}C{1.25cm}@{} @{}C{1.25cm}@{} @{}C{1.25cm}@{} @{}C{1.15cm}@{}}
        \textbf{Model} & {\scriptsize\textbf{BLiMP}} & {\scriptsize\textbf{GLUE}} & {\scriptsize\textbf{MSGS}} & \textbf{Avg.} \\\midrule
        \modelName {\scriptsize\textcolor{gray}{(\textsc{20e})}} & {73.5} & {72.3} & {79.2} & {75.0} \\[2pt]\arrayrulecolor[rgb]{0.753,0.753,0.753}\cline{1-5}
        \hspace{0.2cm}$\neg$ ATF {\scriptsize\textcolor{gray}{(\S\ref{sec:atf})}} & 73.1 & 70.6 & 80.4 & 74.7 \\
        \hspace{0.2cm}+ \textsc{40e} & {74.8} & 73.4 & {80.7} & {76.3} \\
        \hspace{0.2cm}+ \textsc{100e} & {76.5} & {73.8} & {80.0} & {76.8} \\
        \hspace{0.2cm}+ \textsc{200e} & \textbf{76.6} & \textbf{74.0} & 80.9 & \textbf{77.2} \\
        \hspace{0.2cm}+ FLOTA & \textcolor{red}{57.8} & -- & -- & -- \\
        \hspace{0.2cm}+ BRAK {\scriptsize\textcolor{gray}{(40\textsc{e}, \S \ref{subsec:brack})}} & 75.0 & 72.0 & \textbf{82.1} & 76.4 \\[2pt]\arrayrulecolor[rgb]{0.753,0.753,0.753}\cline{1-5}
        dGPT-2 {\scriptsize\textcolor{gray}{($\neg$ATF, 40\textsc{e})}} & 68.9 & 70.2 & 79.9 & 73.0 \\
        \hspace{0.2cm}+ OMG {\scriptsize\textcolor{gray}{(\S \ref{subsec:emergent})}} & 70.8 & 69.7 & 80.0 & 73.5 \\[2pt]\arrayrulecolor[rgb]{0.753,0.753,0.753}\cline{1-5}
        OPT$^\dagger$ & 62.6 & 63.4 & 79.8 & 68.6\\
        RoBERTa$^\dagger$ & 69.5 & 71.4 & 80.9 & 73.9
        \\\arrayrulecolor[rgb]{0,0,0}\bottomrule
    \end{tabular*}
    }
    \caption{
    Aggregate results for the \modelName model with various configurations on the three evaluation suites.
    $n${\sf\textsc{e}} denotes a model trained for $n$ epochs.
    $\dagger$ models are baseline models made available by the BabyLM organisers.
    Best performing model per suite is in \textbf{bold}.
    }
    \label{tab:main_results}
\end{table}
We report the results of our models in Table \ref{tab:main_results}.
Results are aggregated over individual subtasks in BLiMP, GLUE, and MSGS.
Our final \modelName model, trained for 200 epochs with ATF data augmentation, obtained an average score of 77.2.
Next to this model we report various alterations to the training regime.
To investigate the impact of the ATF procedure, we also train a model without the augmented data.
The strongest gains of ATF are achieved in the GLUE tasks (+1.7 points), which is in line with our original goal of aligning the pre-training data more with that of the fine-tuning tasks.
Furthermore, prolonging model training has a strong positive impact on both BLiMP and GLUE, but not for the MSGS tasks. In Figure \ref{fig:blimp} we present a more fine-grained overview of the results split out for each individual task in the evaluation suites, for a subset of models that showcase improvements driven by ATF and prolonged model training.

\paragraph{BLiMP} For BLiMP, increasing the amount of epochs has a positive effect on almost all tasks.
One clear outlier, however, is the Irregular Forms tasks, where our 200 epoch model performs significantly worse than models trained for shorter.
We plot this behavior for models trained on increasing amounts of epochs in Figure \ref{fig:blimp}B, from which it can be seen that this task follows a peculiar \textit{inverse scaling} pattern \citep{mckenzie2023inverse}.
Exploring this pattern in more detail could provide an interesting direction for future research, connecting it to the rule learning of irregular forms in LMs \citep{dankers-etal-2021-generalising}.

\paragraph{GLUE}
The impact of training longer is less pronounced on GLUE than for BLiMP, but it still has a positive effect for most tasks.
The ATF procedure appears to have a positive effect on only a small number of tasks, especially MultiRC and MRPC.
Surprisingly, performance on QNLI and SST2, the tasks targeted by ATF, did not improve significantly.

\section{Additional Experiments}\label{sec:did_not_work}
Our final \modelName model adopted only a small number of the techniques we investigated for the BabyLM challenge.
In this section we highlight various approaches that were not entirely successful, but could serve as inspiration for future work.
Note that some of these approaches would not be permitted under the \texttt{strict-small} conditions of the BabyLM challenge, but would be possible within the \texttt{loose} track.

\subsection{Model Architecture}
\paragraph{FLOTA} 
Our \modelName model uses a BPE subword tokenizer, a common tokenizer used by many LMs, such as GPT-3 \citep{NEURIPS2020_1457c0d6}.
From a linguistic point of view, this tokenization procedure may be sub-optimal: it is based solely on frequency statistics, and takes no morphological information into account (e.g. \textit{undesirable} \textrightarrow~\textit{undesi}+\textit{rable}). 
The FLOTA tokenizer \citep{hofmann-etal-2022-embarrassingly} addresses this concern, and presents a tokenization procedure that adheres more strongly to the morphological formation of English words (e.g. \textit{undesirable} \textrightarrow~\textit{un}+\textit{desirable}).
We incorporated this tokenizer in our pipeline, but unfortunately it resulted in sub-par results on BLiMP (Table \ref{tab:main_results}).
A reason for this might be the relatively low vocabulary size (10.000), though it remains surprising that this tokenizer led to such a significant drop in performance.

\paragraph{LLaMA} LLaMA \citep{touvron2023llama} is a pre-trained model whose performance rivals that of many larger models trained on more data. In order to achieve this performance, it incorporates a variety of architectural tweaks that aim to improve performance or training stability; these include pre-normalization of transformer block inputs using RMSNorm \citep{zhang2019root}, the SwiGLU activation function \citep{shazeer2020glu}, and rotary embeddings \citep{su2022roformer}. Unlike our ChapGTP model, LLaMA uses the SentencePiece tokenizer \citep{kudo-richardson-2018-sentencepiece}. 

Motivated by LLaMA's successful training on smaller data using a smarter architecture, we trained our own LLaMA model. We used a variety of scaled-down model architectures, e.g. with a hidden (residual stream) size of 64, an intermediate (MLP) size of 256, 4 layers, 4 attention heads, and a vocabulary size of 10000. However, these models exhibited no performance gains over similarly sized models that used a more traditional, GPT-like architecture.

\subsection{Model Training}

\paragraph{Prosodic Guidance}
Information in speech is not only conveyed through which words are said, but also how they are spoken \citep{wallbridge_quantifying_2023}. Hence even models trained on transcribed speech data miss out on the rich auditory cues available in spoken language, which could be informative for learning \citep{chrupala_putting_2023}. We explored the use prosodic information as one such guiding signal for language model training.
Prosody is thought to play an important role in scaffolding human language learning \citep{gervain_prosodic_2020}, for example in helping infants learn non-adjacent dependencies by highlighting the relevant linguistic elements \citep{martinez-alvarez_prosodic_2023}. 

One way to provide a text-based language model with a similar learning signal would be to train the model on spoken language transcriptions for which audio recordings are available. Prosodic prominence cues based on properties like pitch and duration, or more advanced scores estimated based on continuous wavelet transforms \citep{suni_hierarchical_2017}, could be extracted from the audio recordings to guide model training. Though we considered this a promising approach to study if language modeling can be improved with access to prosodic information, it was not feasible for us to pursue within the constraints of the BabyLM challenge---curating an audio-aligned text dataset at the 10M- or 100M-word scale poses a significant challenge on its own. We therefore left experiments into using prosodic information for language model training out of our BabyLM submission and hope to work on this idea separately in the future.


\paragraph{Grokking}
Grokking is a phenomenon in which models seemingly neural networks begin to generalize better after overfitting \citep{power2022grokking}. In such scenarios, models initially achieve high training performance, but poor held-out (evaluation) performance. Extended training leads models to suddenly generalize, achieving higher evaluation performance. Grokking has been shown to occur not only on toy algorithmic tasks, but also image and sentiment classification \citep{liu2022towards,liu2023omnigrok}. More recent work has suggested that transformers can grok hierarchical linguistic structure after extremely prolonged training \citep{murty-etal-2023-grokking}.

On the basis of this recent evidence, we conduct experiments to determine if longer training can help language models capture the hierarchical structure of language, even when trained on small data. Our grokking setup is simple: we train a \texttt{DistilGPT2} model for 500 epochs on the small (10M word) dataset. We set training hyperparameters as in \citet{murty-etal-2023-grokking}. We find that grokking does not occur in this scenario: evaluation loss does not improve. Moreover, while our long-training model performed reasonably well on the zero-shot linguistic tasks from BLiMP, performance on the SuperGLUE tasks, which required fine-tuning, is much worse. We conclude that while longer training may not have hurt linguistic knowledge, it may have hurt the model's ability to be fine-tuned. 

These results may be surprising, given that in \S\ref{sec:results}, longer training generally led to better performance on BLiMP and GLUE. Unfortunately, differences in model architecture and training procedure (particularly ATF) could have led to different training dynamics, making direct comparison difficult. Moreover, prior work suggests that the occurrence of grokking is reliant on specific conditions such as a large initial weight norm, or specific adaptive optimizers \citep{thilak2022slingshot,liu2023omnigrok}. More controlled and extensive study is needed to shed light on grokking in LMs.


\subsection{OMG: Data from Object Mediated Games}
\label{subsec:emergent}
Simulating cooperative games with deep neural agents 
that need to communicate about objects in their environment is an active area of research; the communication protocols emerging in these settings have been studied extensively in previous works~\cite[][i.a.]{havrylov2017emergence,kottur2017natural,bouchacourt2018agents,lazaridou2020emergent,luna2020internal}.

An important motivation for these experiments is to simulate conditions under which certain natural language properties may develop~\cite[e.g.][]{kirby2002natural,kirby_compression_2015}. Others suggest that these settings may enable language models to learn aspects of human communication difficult to acquire from passive language modeling alone~\cite[e.g.][]{lazaridou2020multi}.

Interestingly, \citet{yao2022linking} show that pre-pre-training LMs on synthetic emergent languages generated in referential games with images can in fact improve their performance in low-resource settings. We aim to reproduce the findings of \citeauthor{yao2022linking} with our particular setup; as such, compare the performance of \texttt{DistilGPT2} trained on BabyLM with and without first pre-pre-training on their synthetic emergent languages.

\paragraph{Approach}
We pre-pre-train \texttt{DistilGPT2} on a synthetic emergent language coming from a referential game played with neural agents, as provided by \citet{yao2022linking}.\footnote{\url{https://github.com/ysymyth/ec-nl/}}
In this referential game, deep neural agents successfully communicate about images from the Conceptual Captions dataset~\cite{sharma2018conceptual}.
We use the set of messages with vocabulary size 4035 and maximum message length 15, sampling $2,721,927$ messages for the training data, and $143,260$ for the development set (split in roughly $95\%$ and $5\%$, respectively).\footnote{An example of an emergent message (before tokenization and converting to integers) is: \texttt{1019 3876 601 2194 3360 3360 3360 3360 3360 3360 3360 3360 3360 3360 0}.}

We pre-pre-train on the emergent language for 8 epochs, after which we continue pre-training on the BabyLM 10M dataset ($\neg$ ATF) for $40$ epochs.
We compare this to the baseline where we do not pre-pre-train \texttt{DistilGPT2} on the emergent messages.\footnote{Note that the results for this baseline are slightly different from Table \ref{tab:main_results} but comparable, as we used another random seed for training the 40\textsc{e} DistilGPT2.}

\paragraph{Results}
Table \ref{tab:omg_brak} shows the aggregate results of pre-pre-training on synthetic emergent languages (OMG).
Curiously, OMG pre-pre-training seems to result in a better performance on BLiMP and GLUE compared to the baseline.
In our experiments, we also noticed that the loss curves converge faster during training, indicating that OMG pre-pre-training may be a viable strategy for initializing language models in low-resource settings; this is in line with the findings of the original authors~\cite{yao2022linking}.

\begin{table}
    {\sf\footnotesize
    \centering
    \setlength\extrarowheight{3pt}
    \begin{tabular*}{7.7cm}{p{2cm}@{} @{}C{1.25cm}@{} @{}R{1.25cm}@{} @{}C{1.25cm}@{} @{}R{1.25cm}@{}}
        \textbf{Task} & {\scriptsize\textbf{OMG}} & {\scriptsize\textbf{$\Delta_{baseline}$}} & {\scriptsize\textbf{BRAK}} & {\scriptsize\textbf{$\Delta_{baseline}$}} \\\midrule
        BLiMP & 70.8 & +0.7 & 75.0 & -0.6 \\
        GLUE  & 69.7 & +1.4 & 72.0  & -0.7 \\
        MSGS  & 80.0 & -0.1 & 82.1 & +3.2 \\
        \bottomrule
    \end{tabular*}
    }
    \caption{
    Aggregate results for pre-pretraining DistilGPT2 with text from object mediated games (OMG) and \modelName with constituency labelled text (BRAK).
    We also show the difference with their respective baselines ($\Delta_{baseline}$) as discussed in  \S\ref{subsec:emergent} and \S\ref{subsec:brack}, where + indicates an improvement.
    All models shown here are further trained on the BabyLM dataset for 40 epochs \emph{without} the ATF data augmentation (\S\ref{sec:atf}).
    }
    \label{tab:omg_brak}
\end{table}

\subsection{BRAK: Bracketed pre-pre-training}
\label{subsec:brack}
Can initially pre-training on texts where the structure is explicitly marked be used to improve the LM's performance later on?
To test this approach, we train the \emph{Deep Inside-Outside Recursive Autoencoders} model \cite[\texttt{DIORA},][]{drozdov2019diora}, to augment a portion of the training data with \emph{bracketing} that indicate the constituents of the sentences.
The general idea is that the bidirectional \modelName can use this extra training signal to quickly learn the syntactic structures of the data---bootstrapping its further language modeling.

\paragraph{Approach}
We pre-pre-train \modelName for 4 epochs on a subset of $15,030$ sentences from the BabyLM 10M dataset, where the constituents of each sentence is marked using the ``['' and ``]'' tokens.\footnote{An example of a constituency-labeled sentence is: \texttt{[ [ [ [ they are ] placed ] into ] [ [ [ one [ character [ and [ it is ] ] ] ] [ [ mostly [ used with ] ] [ east asian ] ] ] fonts ] ]}.}
After this, pre-training continued on the entirety of the unbracketed BabyLM dataset (without ATF) for $40$ epochs.

To obtain the constituents for the $15,030$ sentences, we trained a \texttt{DIORA} model with a hidden dimension of $50$ and batch size of $128$ for a maximum of $5$ epochs.
We initialized its embeddings using \texttt{GloVe}~\cite{pennington2014glove} (embedding size 16) trained on the same corpus as \texttt{DIORA}.
Since \texttt{DIORA} requires sentences as input, we use the dot (``.'') to split the documents in the datasets into individual sentences, which are then split into words using the space token.
We lower-cased each token and removed all punctuation from the sentences. 
This approach is deliberately kept simple to avoid using any techniques requiring non-trivial expert knowledge.
From this set, we labeled $15,030$ sentences with a minimum length of three with the trained \texttt{DIORA} model.
As a baseline, we pre-pre-train \modelName on the same $15,030$ sentences, but without the bracketing.

\paragraph{Results}
The aggregate results of the bracketed pre-pre-training (BRAK) are shown in Table \ref{tab:main_results} and compared to the baseline in Table \ref{tab:omg_brak}.
While BRAK \modelName performs slightly worse on BLiMP and GLUE, it performs considerably better on the MSGS tasks, as seen in Figure \ref{fig:blimp}D.
BRAK's main gains stem from two tasks: `\textit{Main Verb Lexical Control The}', and `\textit{Main Verb Relative Token Position}'.
We encourage future work on how including inductive biases can improve the performance of language models in low-resource settings.



\section{Conclusion}

In this paper, we introduced our submission to the \texttt{strict-small} track of the BabyLM challenge.
\modelName is a DeBERTa-based masked LM, trained for 200 epochs with help of our novel data augmentation technique: Automatic Task Formation (ATF). 
We proposed ATF as a means of creating more task-specific textual formulations based on the existing training data. In particular, we focused on improving representations for question answering and sentiment classification. The idea behind these specific ATF augmentations was that they might lead our model to learn useful representations for the retrieval- and classification-based GLUE tasks during pre-training; such representations could be harder to learn from the primarily spoken language data in the BabyLM \texttt{strict-small} training set alone.

Our results show that the ATF procedure indeed improved performance on GLUE tasks, especially for the paraphrase detection (MRPC) and multi-sentence reading comprehension (MultiRC) subtasks. The QNLI and SST2 tasks targeted by the Sentiment Classification component of ATF did not improve significantly.
Our experiments with prolonged training of \modelName up to 200 epochs resulted in increased performance for most evaluation benchmarks, but we also found \emph{inverse scaling} behavior for the \emph{Irregular Forms} BLiMP task. Based on this result, exploring how prolonged training affects LM's  memorization of linguistic patterns beyond generalizable rules seems an interesting direction for future research.


\modelName outperforms the baseline models provided by the BabyLM challenge, and our ATF augmentation technique proved successful at improving performance on specific targeted tasks. \citet{jia_question_2022} motivated their QA-infused pre-training approach by the intuition that phrase representations should encode all questions that the phrase can answer in context. Such relational information integration might be encouraged by the addition of ATF question-answer pairs in our augmented training data as well, and could potentially result in more human-like encodings of contextual knowledge.

Nevertheless, the performance of \modelName on BabyLM admittedly does not present significant advances in terms of cognitive plausibility. We believe that promising approaches for stimulating more human-like learning in language models incorporate some form of human-like inductive biases in model training. Since humans presumably come to the language learning task from much less of a ``blank slate’’ state than randomly-initialized masked language models, this area leaves much potential for further research.
Our use of unsupervised constituency parsers for BRAK \modelName (\S\ref{subsec:brack}) was an attempt to make use of such inductive biases in the syntactic domain, and resulted in notable performance gains on hierarchical generalization tasks (MSGS), although ideally such biases would be integrated into LMs more holistically.

Finally, \modelName is only trained only on text, while children rely on many other modalities to learn language (e.g. audition and vision).
Although we made efforts to indirectly incorporate multimodal cues through speech prosody and object-mediated referential games, we only scratched the surface of what is possible. 
The BabyLM challenge provided an inspiring start to explore such possibilities, and we hope that our range of experiments presented here will usefully inform future work on data-efficient and cognitively plausible NLP.



\section*{Limitations}
There are various aspects in our setup that could have been addressed more rigorously.
For reproducibility, the number of random seeds should be increased to obtain more robust insights into the impact of various training enhancements.
The optimality of our hyperparameter setup is not guaranteed, a wider hyperparameter search sweep would be necessary for this.


\section*{Acknowledgements}
The authors gratefully acknowledge Sarenne Wallbridge for insightful discussions about prosody-enhanced LM training, and Jelle Zuidema for useful feedback during the project.


\bibliography{anthology,custom}
\bibliographystyle{acl_natbib}

\appendix

\section{Sentiment Tokens}\label{sec:app_sentiment}
To augment the training corpus with sentiment classification we use the following lists of negative and positive tokens.

\noindent \textbf{Negative}: \{\textit{
    "not good",
    "not great",
    "not like",
    "didn't like",
    "not [a-z]+ great",
    "not [a-z]+ good",
    "horrible",
    "terrible",
    "hate",
    "hated",
    "bad",
    "disliked",
    "annoying",
    "frustrating",
    "worst"
}\}

\noindent \textbf{Positive}: \{\textit{
    "loved",
    "not bad",
    "not [a-z]+ bad",
    "great",
    "fantastic",
    "incredible",
    "terrific",
    "gorgeous",
    "enjoyed",
    "enjoy",
    "beautiful"
}\}

\end{document}